\documentclass[9pt,conference]{IEEEtran}
\usepackage{xurl}  
\usepackage{hyperref} 
\usepackage{cite}
\usepackage{amsmath,amssymb,amsfonts}
\usepackage{algorithmic}
\usepackage{graphicx}
\usepackage{textcomp}
\usepackage{xcolor}
\usepackage{todonotes}

\def\BibTeX{{\rm B\kern-.05em{\sc i\kern-.025em b}\kern-.08em
    T\kern-.1667em\lower.7ex\hbox{E}\kern-.125emX}}

\begin{document}

\title{Learning Speech Attribute Representations \\via Multi-Teacher Distillation\\
\thanks{Identify applicable funding agency here. If none, delete this.}
}

\author{\IEEEauthorblockN{Maryem Bouziane}
\IEEEauthorblockA{\textit{LIA} \\
\textit{Avignon Université}\\
Avignon, France \\
0009-0002-6483-7881}
\and
\IEEEauthorblockN{Salima Mdhaffar}
\IEEEauthorblockA{\textit{LIA} \\
\textit{Avignon Université}\\
Avignon, France \\
0000-0002-8472-6890}
\and
\IEEEauthorblockN{Yannick Estève}
\IEEEauthorblockA{\textit{LIA} \\
\textit{Avignon Université}\\
Avignon, France \\
0000-0002-3656-8883}
}

\maketitle

\begin{abstract}
Recent post-training approaches have shown that self-supervised speech encoders can be adapted to learn a single utterance-level speech attribute through knowledge distillation. In this work, we investigate whether a single speech encoder can simultaneously preserve multiple complementary utterance-level speech representations. We propose a multi-attribute multi-teacher distillation framework that learns semantic, speaker, language, and emotion representations from four frozen teacher models using a shared SSL backbone and attribute-specific branches. Experiments on multilingual semantic retrieval, speaker verification, language identification, and speech emotion recognition show that the proposed model achieves performance close to dedicated single-task students while requiring only a single shared encoder. Furthermore, the learned layer aggregation strategies reveal that different speech attributes naturally emerge at different depths of the SSL encoder, providing new insights into the organization of speech representations.
\end{abstract}

\section{Introduction}

Speech conveys multiple complementary types of information beyond its linguistic content. In addition to the lexical message, an utterance carries information about the speaker identity, the spoken language, emotional state, speaking style, and other paralinguistic characteristics. These complementary attributes are essential for a broad range of applications including multilingual speech retrieval, speaker verification, language identification, conversational agents, speech analytics, and affective computing. Nevertheless, they are generally modeled independently, leading to multiple task-specific encoders that duplicate computation and fail to exploit the common structure underlying speech.

Recent advances in self-supervised learning (SSL) have profoundly changed speech representation learning. Foundation models such as wav2vec~2.0~\cite{baevski2020wav2vec}, HuBERT~\cite{hsu2021hubert}, WavLM~\cite{chen2022wavlm}, and w2v-BERT~\cite{chung2021w2v} are trained on thousands of hours of unlabeled speech and produce generic contextual representations that achieve remarkable performance across many downstream tasks. While these models naturally produce frame-level representations, many applications require utterance-level embeddings that summarize global properties of an entire speech segment. Rather than redesigning SSL pre-training objectives, several recent works have demonstrated that these foundation models can be effectively adapted through \emph{post-training}, transferring knowledge from pretrained expert models into a shared speech encoder.

A particularly successful direction consists in learning semantic speech embeddings through teacher-student knowledge distillation. Frameworks such as SAMU-XLSR~\cite{khurana2022samu}, SONAR~\cite{Duquenne:2023:sonar_arxiv,seamless2025joint}, and SENSE~\cite{mdhaffar2025sense} align speech representations with multilingual sentence embedding spaces, enabling multilingual and multimodal retrieval. More recently, multi-teacher post-training has been proposed to jointly learn semantic and speaker representations within a single SSL speech encoder~\cite{bouziane2026multiattr}. These results demonstrate that multiple utterance-level attributes can coexist in a unified representation while preserving competitive performance on both semantic retrieval and speaker verification.

However, extending this paradigm beyond two attributes remains an open challenge. Semantic content, speaker identity, language identity, and emotion emerge at different abstraction levels within SSL speech encoders and rely on heterogeneous supervision signals originating from different datasets and teacher models. Simply adding new objectives may lead to conflicting optimization dynamics and representation interference. Whether a single post-trained SSL encoder can simultaneously preserve these complementary dimensions of speech therefore remains largely unexplored.

In this paper, we address this challenge by proposing a holistic multi-teacher post-training framework that learns four complementary utterance-level representations from a single shared SSL speech encoder. The proposed framework distills semantic, speaker, language, and emotion representations from four specialized frozen teachers while introducing two key mechanisms to facilitate joint optimization. First, each attribute learns its own layer aggregation strategy, allowing different branches to exploit the encoder layers that are most informative for their target representation. Second, we propose a gain-based loss weighting strategy derived from mono-task learning dynamics to balance heterogeneous teachers exhibiting different convergence behaviors.

We evaluate the proposed framework on four downstream tasks covering all distilled attributes: multilingual semantic retrieval, speaker verification, language identification, and speech emotion recognition. 

The contributions of this work are fourfold:
\begin{itemize}
    \item We significantly extend recent multi-teacher post-training approaches by jointly learning four complementary utterance-level speech attributes instead of two.
    \item We propose a holistic post-training framework for SSL speech encoders combining attribute-specific layer aggregation with a gain-based loss weighting strategy.
    \item We demonstrate that a single shared encoder preserves semantic, speaker, language, and emotion representations across four independent downstream evaluation tasks.
    \item We provide an analysis of layer utilization showing how different speech attributes emerge at different depths of a shared SSL speech encoder.
\end{itemize}

\section{Related Work}

\subsection{Post-training of speech foundation models}


Recent work has increasingly viewed post-training as a distinct stage in the lifecycle of speech foundation models, complementing large-scale self-supervised pre-training. Rather than modifying the expensive pre-training process, post-training adapts a generic speech encoder to downstream objectives through techniques including domain adaptation, instruction tuning, continual learning, and knowledge distillation \cite{li2023efficient,ge2025post}.
Existing post-training approaches can be broadly categorized into four directions: adaptation, alignment, reasoning, and efficiency \cite{ptsfm2026}. Such post-training approaches have also explored efficiency through model compression and knowledge distillation~\cite{hinton2015distilling}. 
In this work, rather than compressing a single expert, we consolidate the knowledge of multiple specialized teacher models into a single shared speech encoder, reducing the number of deployed encoders while preserving multiple speech attributes. 
The resulting model replaces multiple task-specific encoders with a single backbone capable of producing semantic, speaker, language, and emotion representations.

\subsection{Learning utterance-level speech representations}

While self-supervised speech encoders naturally produce contextual frame-level representations, many downstream applications require compact utterance-level embeddings that summarize an entire speech segment. 
Early efforts mainly focused on learning a single attribute. Semantic speech representations have received particular attention through teacher-student distillation, where speech utterances are aligned with multilingual sentence embedding spaces. 
Representative examples include SAMU-XLSR~\cite{khurana2022samu}, SONAR~\cite{Duquenne:2023:sonar_arxiv,seamless2025joint}, and SENSE~\cite{mdhaffar2025sense}, which produce language-agnostic utterance embeddings suitable for multilingual speech-speech and speech-text retrieval. 
These approaches demonstrate that semantic information can be effectively transferred from powerful text encoders to speech foundation models while preserving cross-lingual generalization.

Beyond semantics, several studies have investigated utterance-level representations dedicated to specific paralinguistic attributes. Speaker verification systems typically rely on speaker embeddings extracted from models such as ECAPA-TDNN~\cite{desplanques2020ecapa}, while language identification has similarly benefited from utterance-level embeddings trained on multilingual corpora such as VoxLingua107~\cite{valk2021voxlingua107}. 
More recently, speech emotion representation learning has attracted increasing attention, with models such as emotion2vec~\cite{ma2024emotion2vec} demonstrating that large-scale self-supervised pre-training can produce transferable emotion embeddings across a wide range of downstream emotion recognition tasks.

Although these representations target different speech properties, they share a common characteristic: each embedding space is optimized independently for a single attribute. 
As a consequence, practical systems typically require several specialized encoders operating in parallel, increasing memory usage, computational cost, and deployment complexity. 
Only recently have a few studies begun to investigate whether multiple utterance-level representations can be jointly learned from a shared speech encoder through multi-teacher distillation~\cite{bouziane2026multiattr}. 
These first results suggest that multiple speech attributes can coexist within a common SSL encoder, opening new perspectives for unified utterance-level representation learning.

\section{Method}

\begin{figure*}[!t]
\centerline{\includegraphics[width=2\columnwidth]{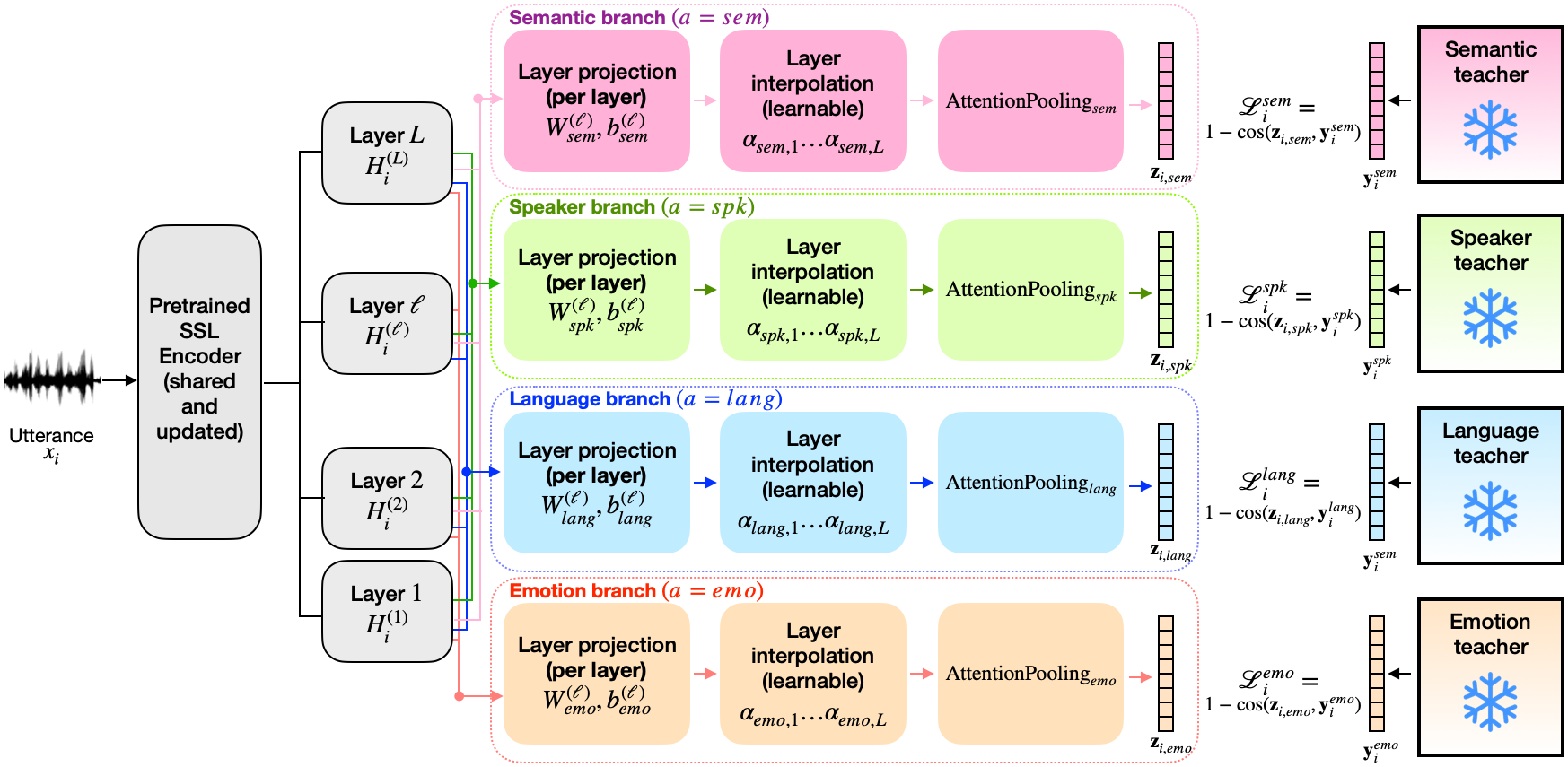}}
\caption{Overview of the multi-teacher distillation architecture.}
\label{fig:simplifed_archi_multi_attr}
\end{figure*}

Figure~\ref{fig:simplifed_archi_multi_attr} illustrates the proposed framework. A shared SSL speech encoder processes the input utterance and produces contextual hidden representations at all encoder layers. 
Four attribute-specific branches, corresponding to semantic content, speaker identity, language identity and emotion, independently aggregate these hidden representations through learnable layer interpolation, followed by attention pooling and a projection layer. Each branch is supervised by a frozen teacher model through cosine embedding distillation. 
The entire architecture is trained end-to-end while sharing a single speech encoder across all attributes.

We propose a holistic multi-teacher distillation framework for learning several utterance-level speech representations with a single shared speech encoder. 
Let $\mathcal{A}$ denote the set of target attributes:
\[
\mathcal{A} = \{\mathrm{sem}, \mathrm{spk}, \mathrm{lang}, \mathrm{emo}\},
\]
corresponding respectively to semantic content, speaker identity, language identity, and emotion. 
Each attribute $a \in \mathcal{A}$ is associated with a frozen teacher model $T_a$, which defines a target embedding space for that attribute. The student model consists of a shared speech foundation encoder followed by attribute-specific branches. 
Each branch produces one utterance-level embedding and is trained to match the corresponding teacher representation.
Given a speech utterance $x_i$, each teacher produces a target embedding $\mathbf{y}_{i}^{a} = T_a(x_i)$.

\subsection{Shared encoder with attribute-specific branches}

The student encoder is initialized from a pretrained speech foundation model. 
Given an input utterance $x_i$, the encoder produces a sequence of hidden representations at different layers:
\[
\mathbf{H}_{i}^{(\ell)} \in \mathbb{R}^{T_i \times D},
\]
where $T_i$ is the number of frames, $D$ is the dimension of hidden representations, and $\ell \in \{1,\ldots,L\}$ denotes the encoder layer. 
Instead of using a fixed hidden layer for all attributes, we allow each attribute-specific branch to learn which encoder layers are most useful for its corresponding teacher.


For each attribute $a$, we attach a dedicated branch to the shared encoder. First, each hidden layer is projected into an attribute-specific feature space:

\[
\widetilde{\mathbf{H}}_{i,a}^{(\ell)}
=
\mathbf{H}_{i}^{(\ell)} \mathbf{W}_{a}^{(\ell)\top}
+
\mathbf{b}_{a}^{(\ell)}.
\]
where the projection maps the shared hidden dimension into an attribute-specific feature space.

Different speech attributes are known to emerge at different depths of SSL encoders. Lower layers mainly capture acoustic information, whereas deeper layers encode increasingly semantic information. Rather than selecting a fixed encoder layer for all attributes, we allow each branch to learn an attribute-specific interpolation over all hidden layers.
To model the fact that different attributes may rely on different levels of abstraction, each branch learns a scalar score $s_{a,\ell}$ for every encoder layer. These scores are converted into normalized interpolation weights using a softmax:
\[
\alpha_{a,\ell}
=
\frac{\exp(s_{a,\ell})}
{\sum_{j=1}^{L} \exp(s_{a,j})},
\qquad
\sum_{\ell=1}^{L} \alpha_{a,\ell}=1.
\]

The projected layer representations are then combined as:
\[
\widehat{\mathbf{Z}}_{i,a}
=
\sum_{\ell=1}^{L}
\alpha_{a,\ell}
\widetilde{\mathbf{H}}_{i,a}^{(\ell)}.
\]
Then a layer normalization is applied to obtain $\mathbf{Z}_{i,a}$

The resulting frame-level sequence is converted into an utterance-level representation using an attribute-specific single-head self-attention module~\cite{safari2020self}:
\[
\mathbf{p}_{i,a}
=
\mathrm{AttentionPooling}_{a}(\mathbf{Z}_{i,a}).
\]

Finally, an optional projection maps the pooled representation in order to fit to the teacher embedding dimension:
\[
\mathbf{z}_{i,a}
=
\mathrm{Proj}_{a}(\mathbf{p}_{i,a}).
\]
Both the student output $\mathbf{z}_{i,a}$ and the teacher target $\mathbf{y}_{i}^{a}$ are $\ell_2$-normalized before computing the distillation loss.

\subsection{Multi-teacher distillation objective}

For each attribute $a$, the student branch is trained to maximize the cosine similarity with the corresponding teacher representation. The attribute-specific loss is defined as:
\[
\mathcal{L}_{i}^{a}
=
1 -
\cos(\mathbf{z}_{i,a}, \mathbf{y}_{i}^{a}).
\]


A central difficulty in holistic multi-teacher distillation is that the different teachers have heterogeneous embedding dimensions, training dynamics, and levels of difficulty. 
Uniform loss weighting may therefore over-emphasize attributes that are easier to optimize or whose losses have larger magnitudes.

Mono-task learning provides an estimate of how much useful information can be transferred from each teacher into the student architecture.
Attributes exhibiting larger learning gains tend to correspond to teachers that are easier for the student to approximate.

Assigning them larger weights encourages the Multi-Attribute optimization to focus on supervision signals that are better aligned with the student capacity, while avoiding over-emphasizing attributes that are intrinsically difficult or noisy.

We introduce a gain-based weighting strategy derived from mono-task learning dynamics. 
For each attribute $a$, we first train a mono-task student using the same architecture but retaining only branch $a$. 
Let $\mathcal{L}_{a}^{\mathrm{mono}}(e)$ denote the normalized validation loss of this mono-task model at epoch $e$. 
The learning gain for attribute $a$ is defined as:
\[
g_a
=
\mathcal{L}_{a}^{\mathrm{mono}}(1)
-
\min_e \mathcal{L}_{a}^{\mathrm{mono}}(e).
\]

The final Multi-Attribute weight is obtained by normalizing the gains across attributes:
\[
\lambda_a
=
\frac{g_a}
{\sum_{b \in \mathcal{A}} g_b}.
\]

This strategy assigns larger weights to attributes for which the student shows stronger learning progress in the corresponding mono-task setting, while reducing the contribution of attributes that rapidly saturate or provide weaker training gains. 
Unlike dynamic weighting strategies such as uncertainty weighting or GradNorm~\cite{chen2018gradnorm}, the proposed weights are estimated once from mono-task learning dynamics and remain fixed during multi-teacher training, avoiding additional optimization variables and training instability.

For a training utterance $x_i$, the overall distillation loss is defined as

\begin{equation}
\mathcal{L}_i
=
\sum_{a \in \mathcal{A}}
\lambda_a
\mathcal{L}_i^{a},
\label{eq:global_loss}
\end{equation}

Although four attributes are considered in this work, the proposed framework is generic. Adding a new utterance-level attribute only requires a frozen teacher model and an additional attribute-specific branch, without modifying the shared speech encoder or the optimization procedure.

\section{Experiments}

In order to evaluate the relevance of our proposition, we benchmark it on different tasks related to the targeted sentence-level attributes.

\subsection{Teacher supervision}


We target four utterance-level attributes in our multi-teacher student learning framework: semantics, speaker, language, and emotion continuous representations.
For each attribute, we used a specialized teacher:  the BGE-M3 text encoder (when transcription is available) or the SENSE speech encoder for semantics (BGE-M3 and SENSE share the same semantic continuous space), an ECAPA-TDNN speaker embeddings trained on VoxCeleb for speaker attribute, an ECAPA-TDNN language embedding model trained on VoxLingua107, and emotion2vec-base~\cite{ma2024emotion2vec} as a teacher for emotion attribute. 
All teacher models remain frozen throughout training.



\subsection{Datasets}

The training data of our sentence-level multiple attribute speech encoder combines Common Voice~19~\cite{ardila2020common} and a collection of dataset annotated with emotion labels in different languages. 
This list of emotion datasets used in our collection is presented in Table~\ref{tab:emo_dataset_stats}.

From Common Voice 19, we use the 83  languages supported by BGE-M3. 
Common Voice is mainly used for semantic, speaker, and language supervision, while the emotion collection provides affective speech data for emotion supervision and additional non-transcribed speech.
For the emotion collection, each corpus is split independently into 80\% training, 10\% development, and 10\% test data.
After preprocessing, Common Voice provides 8,205.92 hours of speech and the emotion corpora provide 92.59 hours for training. 
This results a total of 8,298.51 hours of training data. 
The emotion collection contains 31 speech emotion recognition datasets covering multiple languages, recording conditions, and emotion taxonomies. 
Table~\ref{tab:emo_dataset_stats} reports the language, number of emotion classes, and duration of each
emotion dataset.

\begin{table}[htbp]
\caption{Emotion datasets used for emotion supervision.}
\begin{center}
\footnotesize
\setlength{\tabcolsep}{2.5pt}
\begin{tabular}{|l|c|c|c|}
\hline
\textbf{Dataset} & \textbf{Language} & \textbf{\# Classes} & \textbf{Hours} \\
\hline
ESD~\cite{zhou2021seen} & Mixed & 5 & 23.23 \\
\hline
ASVP-ESD~\cite{dejoli2021audio} & Mixed & 12 & 11.35 \\
\hline
EmoV-DB~\cite{adigwe2018emotional} & English & 5 & 7.53 \\
\hline
MELD~\cite{poria2019meld} & English & 7 & 6.36 \\
\hline
PAVOQUE~\cite{steiner2013pavoque} & German & 5 & 6.28 \\
\hline
SUBESCO~\cite{sultana2021sust}& Bangla & 7 & 6.26 \\
\hline
Emozionalmente~\cite{catania2025emozionalmente} & Italian & 7 & 5.04 \\
\hline
IEMOCAP~\cite{busso2008iemocap} & English & 4 & 4.50 \\
\hline
CREMA-D~\cite{cao2014crema}& English & 6 & 4.21 \\
\hline
nEMO~\cite{christop2024nemo} & Polish & 6 & 2.46 \\
\hline
ShEMO~\cite{nezami2019shemo} & Persian & 6 & 2.18 \\
\hline
ASED~\cite{retta2023new} & Amharic & 5 & 1.69 \\
\hline
EMNS~\cite{noriy2023emnsimzcorpusemotive}& English & 8 & 1.45 \\
\hline
TESS~\cite{ma2024emobox}& English & 7 & 1.28 \\
\hline
RAVDESS~\cite{livingstone2018ryerson}& English & 8 & 1.19 \\
\hline
RESD~\cite{ma2024emobox} & Russian & 7 & 1.17 \\
\hline
JL-Corpus~\cite{james2018open}& English & 5 & 1.13 \\
\hline
CaFE~\cite{gournay2018canadian}& French & 7 & 0.86 \\
\hline
eNTERFACE~\cite{martin2006enterface}& English & 6 & 0.81 \\
\hline
AESDD~\cite{vryzas2018speech} & Greek & 5 & 0.55 \\
\hline
CASIA~\cite{ma2024emobox} & Mandarin & 6 & 0.51 \\
\hline
SAVEE~\cite{jackson2014surrey} & English & 7 & 0.41 \\
\hline
EMOVO~\cite{costantini2014emovo} & Italian & 7 & 0.40 \\
\hline
TurEV-DB~\cite{canpolat2020turkish} & Turkish & 4 & 0.37 \\
\hline
EmoDB~\cite{burkhardt2000database} & German & 7 & 0.33 \\
\hline
ANAD~\cite{alnuaim2022human} & Arabic & 3 & 0.27 \\
\hline
OREAU~\cite{ma2024emobox} & French & 7 & 0.27 \\
\hline
URDU~\cite{latif2018cross} & Urdu & 4 & 0.22 \\
\hline
MESD~\cite{duville2021mexican} & Spanish & 6 & 0.14 \\
\hline
Polish~\cite{ma2024emobox} & Polish & 3 & 0.08 \\
\hline
Hi, KIA~\cite{kim2022hi} & Korean & 4 & 0.07 \\
\hline
\textbf{Total} & -- & -- & \textbf{92.59} \\
\hline
\end{tabular}
\label{tab:emo_dataset_stats}
\end{center}
\end{table}


\subsection{Sampling strategy}
In order to balance the data sources across tasks and languages, we applied a resampling strategy similar to those proposed in~\cite{khurana2022samu} and~\cite{mdhaffar2025sense}, which were used to train the SAMU-XLSR and SENSE models, respectively. 
This strategy computes a sampling weight for each sample. 
These weights are then used by the sampler to randomly select samples when constructing each training batch.

\subsection{Implementation and training details}
The shared speech encoder is initialized from w2v-BERT 2.0. 
We implement the multi-teacher model in SpeechBrain, based on the open-source SENSE implementation. 
All teacher embeddings are pre-computed before training
in order to reduce the training cost.
We fine-tune the shared w2v-BERT 2.0 encoder with Adam using a learning rate of $10^{-5}$. 
The task-specific modules are optimized separately
with Adadelta using an initial learning rate of 1.5. 
Training is performed with a batch size of 14 for both training and validation. 
The model is trained for 17,613 iterations using 32 H100 GPUs.
\subsection{Tasks}
\label{sec:evaluation}


To evaluate the quality of the semantic representation learned by our approach, we assess the model on multilingual speech retrieval. 
We evaluate the learned speaker
representations on a speaker verification benchmark, and the language branch through a language identification experiment based on a classifier trained on the extracted embeddings.
Last, we evaluate the emotion branch on speech emotion
recognition datasets using dataset-specific classifiers.

For each attribute, we compare three types of representations: the
corresponding frozen teacher model, a single-task student model trained
only for the considered attribute, and the proposed multi-teacher
student model. This comparison allows us to quantify the impact of
joint training and to evaluate whether a unified encoder can preserve
attribute-specific information while sharing most of its parameters
across tasks.

\subsubsection{Multilingual speech retrieval}

We assess the semantic branch on multilingual speech retrieval. 
For each experiment, we define a query set and a search set. 
The query set contains speech utterances, while the search set contains the corresponding translated
items in another language. 
The objective is to retrieve the correct translation of each query from all candidates in the search set.
Each query utterance is encoded into an utterance-level embedding using the semantic branch of the evaluated model. 
When the search set contains speech, candidate utterances are encoded with the same speech encoder.
When the search set contains text, target sentences are encoded with BGE-M3.
Embeddings are normalized before retrieval, and candidates are ranked according to cosine similarity. Performance is reported with Recall@1.
In this work, we restrict the semantic evaluation to VoxPopuli and FLEURS. 
VoxPopuli is used to evaluate multilingual speech retrieval on European languages, while FLEURS is used to assess generalization to a broader multilingual setting, including unseen and low-resource languages. 
We compare the semantic branch of the proposed multi-teacher model with the semantic-only student and the SENSE teacher.

\subsubsection{Speaker verification}

We evaluate the quality of speaker-related information using the VoxCeleb1-O speaker verification protocol~\cite{nagrani2020voxceleb}. 
Speaker embeddings are extracted from enrolment and test utterances using the speaker branch of the evaluated student models, or directly using the ECAPA-TDNN speaker teacher.
For each trial pair, verification scores are computed using cosine similarity between the two utterance-level speaker embeddings. 
We compare the proposed multi-teacher model with the frozen ECAPA-TDNN speaker teacher and with a single-task speaker student trained using the same architecture but retaining only the speaker branch. 
Performance
is reported in terms of Equal Error Rate (EER) and minimum normalized Detection Cost Function (minDCF), with $P_{\mathrm{target}}=0.01$ and $C_{\mathrm{FA}}=C_{\mathrm{Miss}}=1$.

\subsubsection{Language identification}
To evaluate whether the language branch captures language-discriminative
information, we perform a language identification experiment on
VoxLingua107. For each model, we first extract utterance-level language
embeddings using the corresponding language representation: the frozen
language teacher, the single-task language student, or the language
branch of the multi-teacher student.
A lightweight classifier is trained on top of the extracted embeddings
using the VoxLingua107 training set and evaluated on its development
set. The embedding extractor remains frozen
during this classifier training stage. This protocol measures the
quality of the learned language embedding space independently of the
multi-teacher training objective.
Performance is reported in terms of classification error rate. A lower
error rate indicates that the extracted embeddings contain more
language-discriminative information.
\subsubsection{Speech emotion recognition}

We evaluate the emotion branch on speech emotion recognition datasets.
For each emotion dataset, we extract utterance-level emotion embeddings
from the train, validation, and test partitions using three
representations: the frozen emotion teacher, the single-task emotion
student, and the emotion branch of the multi-teacher student.
For each dataset and each representation type, we train a separate
emotion classifier on the training partition of that dataset. The
validation partition is used for model selection, and the final
performance is reported on the test partition. This dataset-specific
evaluation protocol avoids mixing incompatible emotion taxonomies across
corpora and allows each classifier to match the number of emotion
classes defined for the corresponding dataset.
The embedding extractor remains frozen during classifier training. The classifier is therefore used only as a probing model to assess how much emotion-discriminative information is contained in the extracted representations. Performance is reported using unweighted accuracy (UA), which gives equal importance to each emotion class and is therefore suitable for evaluating emotion recognition performance under class imbalance.

\subsection{Experimental results}
\label{sec:results}

In the table of results, we use the following notation. SENSE denotes the semantic teacher, ECAPA-spk denotes the speaker teacher, ECAPA-lang denotes the language teacher, and emotion2vec-base denotes the emotion teacher. 
SEM-only, SPK-only, LANG-only, and EMO-only refer to single-task students trained with the same architecture as the proposed model but optimized with only one attribute-specific objective. 
Multi-Attribute denotes our proposed model trained jointly with semantic, speaker, language, and emotion supervision.

\subsubsection{Semantic attribute}
Tables~\ref{tab:voxpopuli_all} and~\ref{tab:fleurs_results} report Recall@1 scores for cross-lingual speech-to-speech retrieval on VoxPopuli and FLEURS corpora. 
Overall, the Multi-Attribute model preserves semantic retrieval performance very well and often improves over the SEM-only student or the SENSE teacher. 
These results indicate that Multi-Attribute supervision does not degrade the semantic space; on the contrary, the additional objectives can act as a regularizer and improve semantic retrieval robustness.


\begin{table*}[htbp]
\centering
\caption{Recall@1 scores (\%) for speech-to-speech retrieval on VoxPopuli.}
\label{tab:voxpopuli_all}
\setlength{\tabcolsep}{1.2pt}
\renewcommand{\arraystretch}{1.1}
\begin{tabular}{|c|ccccccc|ccccccc|cccccccc|}
\hline
\textbf{Model}
& \multicolumn{7}{c|}{\textbf{X $\rightarrow$ EN}}
& \multicolumn{7}{c|}{\textbf{EN $\rightarrow$ Y}}
& \multicolumn{8}{c|}{\textbf{X $\rightarrow$ Y}} \\
\cline{2-23}
& \textbf{fr-en} & \textbf{pl-en} & \textbf{nl-en} & \textbf{es-en}
& \textbf{hr-en} & \textbf{de-en} & \textbf{ro-en}
& \textbf{en-fr} & \textbf{en-pl} & \textbf{en-nl} & \textbf{en-es}
& \textbf{en-hr} & \textbf{en-de} & \textbf{en-ro}
& \textbf{fr-de} & \textbf{hr-cs} & \textbf{ro-fr} & \textbf{hu-da}
& \textbf{de-fr} & \textbf{cs-hr} & \textbf{fr-ro} & \textbf{da-hu} \\
\hline
SENSE
& 96.55 & 96.46 & 95.75 & 96.48 & 96.50 & 94.71 & 96.83
& 96.54 & 96.25 & 95.71 & 96.37 & 96.31 & 94.12 & 97.16
& 95.20 & 94.75 & 96.55 & 92.69 & 95.36 & 94.69 & 96.90 & 92.63 \\
\hline
SEM-only
& 94.05 & 93.83 & 92.62 & 94.62 & 95.17 & 91.21 & 95.90
& 93.96 & 94.66 & 92.77 & 94.12 & 95.96 & 90.65 & 96.24
& 90.79 & 94.03 & 95.58 & 89.71 & 91.94 & 93.81 & 96.06 & 90.73 \\
\hline
Multi-Attribute
& 95.19 & 94.85 & 94.55 & 95.45 & 95.41 & 92.68 & 96.10
& 95.10 & 94.85 & 94.62 & 95.30 & 95.52 & 92.29 & 96.37
& 92.43 & 93.63 & 95.75 & 88.41 & 92.31 & 93.63 & 96.07 & 88.46 \\
\hline
\end{tabular}
\vspace{-0.4cm}
\end{table*}



\begin{table}[htbp]
\centering
\caption{Recall@1 scores for multilingual retrieval on FLEURS.}
\label{tab:fleurs_results}

\setlength{\tabcolsep}{2pt}
\renewcommand{\arraystretch}{1.1}

\begin{tabular}{|c|cccccccc|}
\hline
\textbf{Model}
& \textbf{ml-en} & \textbf{lb-en} & \textbf{uz-en} & \textbf{bs-en}
& \textbf{my-en} & \textbf{ny-ces} & \textbf{sd-fr} & \textbf{xh-ar} \\
\hline
SENSE
& 62.59 & 60.88 & 55.40 & 60.39 & 14.11 & 27.00 & 59.74 & 36.67 \\
\hline
SEM-only
& 63.11 & 60.99 & 56.10 & 59.52 & 19.11 & 29.04 & 58.67 & 37.62 \\
\hline
Multi-Attribute
& 62.90 & 60.66 & 56.10 & 59.85 & 20.59 & 25.37 & 57.92 & 36.19 \\
\hline
\end{tabular}

\end{table}

\subsubsection{Speaker verification}
Table~\ref{tab:speaker_results} presents the speaker verification results on VoxCeleb1-O. 
The ECAPA-spk teacher and the SPK-only student obtain the same EER of 0.899\%. 
The Multi-Attribute model obtains an EER of 0.915\%, corresponding to a very small absolute increase of 0.016 points. 
The minDCF values are also close across the three systems. 
These results show that joint Multi-Attribute training introduces only a marginal cost for speaker verification, while allowing the same encoder to support several other utterance-level representations.

\begin{table}[htbp]
\caption{Speaker verification results on VoxCeleb1-O.}
\begin{center}
\begin{tabular}{|c|c|c|}
\hline
\textbf{Model} & \textbf{EER (\%) $\downarrow$} & \textbf{minDCF $\downarrow$} \\
\hline
ECAPA-spk & 0.899 & 0.001104 \\
\hline
SPK-only & 0.899 & 0.001207 \\
\hline
Multi-Attribute & 0.915 & 0.001229 \\
\hline
\end{tabular}
\label{tab:speaker_results}
\end{center}
\end{table}

\subsubsection{Language identification}
Table~\ref{tab:language_results} reports language identification performance. 
The ECAPA-lang teacher obtains the lowest error rate, with 6.58\%. 
LANG-only remains close to the teacher, reaching 6.83\%. 
The Multi-Attribute model obtains 7.02\% with its language attribute representation, corresponding to an absolute increase of only 0.19 points compared with LANG-only and 0.44 points compared with the teacher. Interestingly, by combining the language attribute with the semantic attribute generated by our model, we outperform the results reached by the ECAPA-lang teacher, showing that such a multi-attribute speech encoder provides complementary information useful to process some tasks.


\begin{table}[htbp]
\caption{Language identification results on VoxLingua107.}
\begin{center}
\begin{tabular}{|c|c|}
\hline
\textbf{Model} & \textbf{Classification error (\%) $\downarrow$} \\
\hline
ECAPA-lang teacher & 6.58 \\
\hline
LANG-only & 6.83 \\
\hline
Multi-Attribute Lang & 7.02 \\
\hline
\hline
Multi-Attribute Sem + Lang & \textbf{6.39} \\

\hline
\end{tabular}
\label{tab:language_results}
\end{center}
\end{table}

\subsubsection{Speech emotion recognition}

\begin{figure}[htbp]
\centerline{\includegraphics[width=1.08\columnwidth]{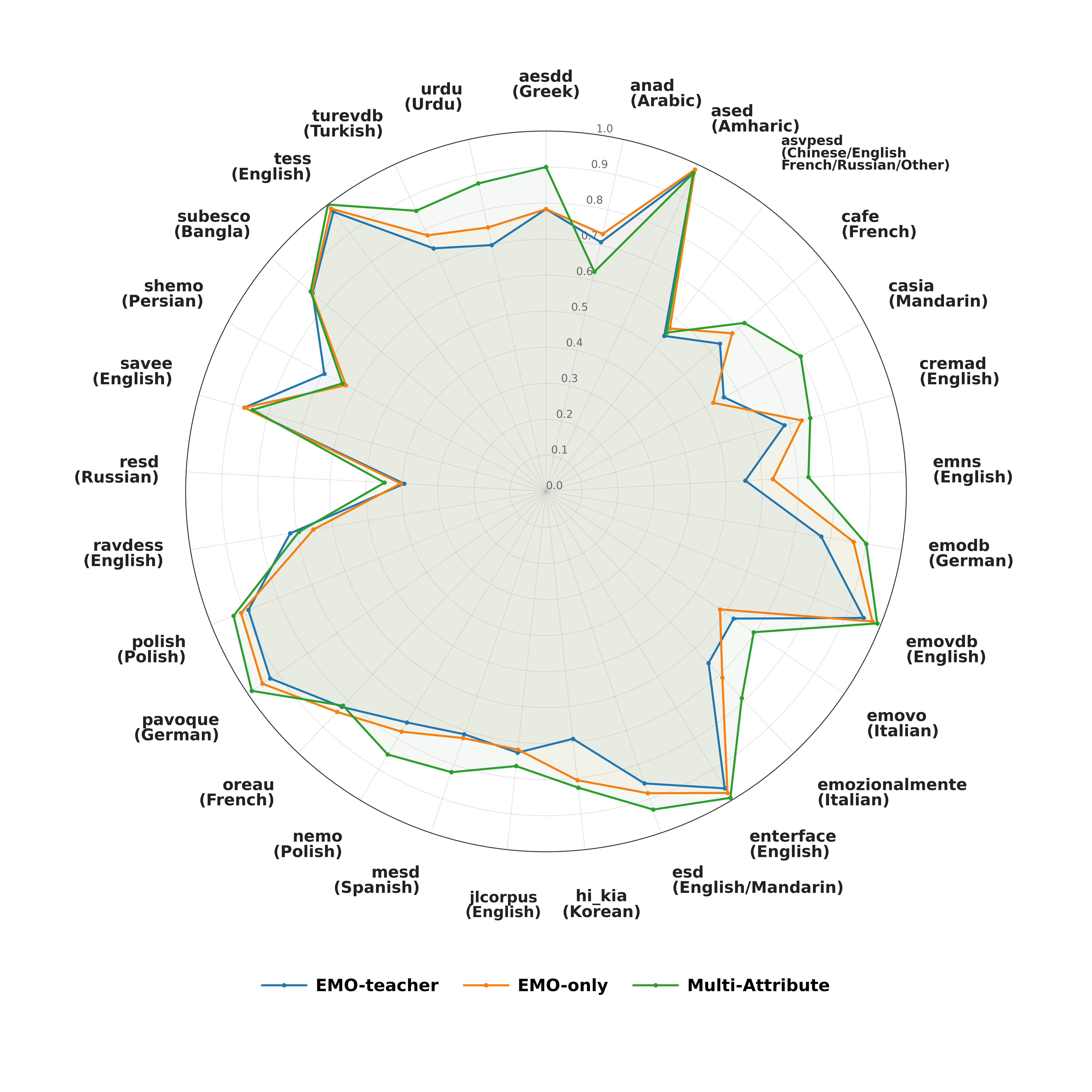}}
\caption{Unweighted Accuracy on speech emotion recognition datasets. The figure compares the emotion2vec-base teacher, the EMO-only student, and the emotion branch of the proposed Multi-Attribute model.}
\label{fig:ser_ua_radar}
\end{figure}

We finally evaluate the emotion branch on speech emotion recognition (SER). 
We compare three representations: the emotion2vec-base teacher, the EMO-only student, and the emotion branch of the Multi-Attribute model.
Figure~\ref{fig:ser_ua_radar} reports Unweighted Accuracy (UA) for each dataset.
The radar plot shows that in many datasets, Multi-Attribute matches or improves over both the EMO-only student and the emotion2vec-base teacher, indicating that the emotion branch benefits from the shared encoder rather than being systematically degraded by the additional objectives. 


Overall, the results across the four evaluation tasks show that the
proposed multi-teacher student can learn a unified speech encoder that
preserves multiple utterance-level attributes simultaneously. Compared
with single-task students, the Multi-Attribute model achieves competitive
performance on speaker verification, language identification, and speech
emotion recognition, while improving semantic retrieval on VoxPopuli and
remaining competitive on FLEURS. These findings support the central
hypothesis of this work: semantic, speaker, language, and emotion
representations can be distilled into a single shared speech encoder with
only limited trade-offs across attributes.

\subsection{Layer-wise analysis}
\label{sec:analysis}

We are also interested by the changes due to the Multi-Attribute learning within the internal representations of the SSL backbone model.
We analyze the learned layer-interpolation weights
$\alpha_{\tau,\ell}$ associated with each attribute-specific branch.
These weights give an indication on how much each encoder layer contributes to the final utterance-level representation of a given attribute $\tau$.

\begin{figure}[!t]
\centerline{\includegraphics[width=0.90\columnwidth]{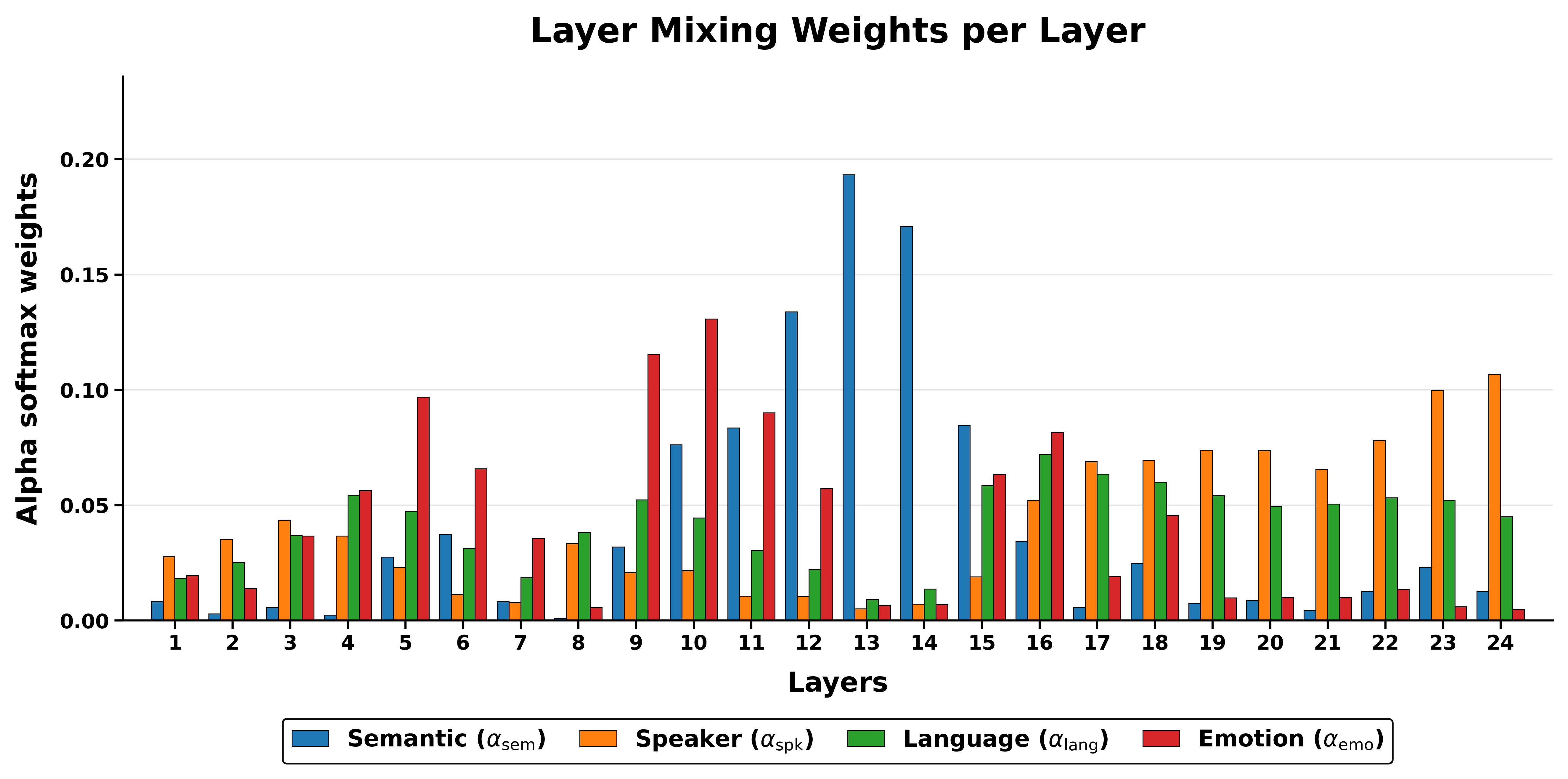}}
\caption{Layer-interpolation weights learned by the four branches of the Multi-Attribute model. Semantic information is mainly extracted from intermediate layers, while speaker and language information are distributed across several encoder layers.}
\label{fig:layer_alphas_multitask}
\end{figure}

Figure~\ref{fig:layer_alphas_multitask} shows the layer weights learned by the Multi-Attribute model. The semantic branch mainly relies on intermediate layers, with the largest weights around layers 12-14. 
This suggests that semantic information is primarily captured by contextual middle-level representations. In contrast, speaker and language information are not concentrated in a narrow set of layers. 
Their weights are distributed across a broader part of the encoder, indicating that speaker- and language-related cues are distributed across several levels of the representation.
The emotion branch follows a different trend, with stronger contributions from early-to-middle layers. 
This suggests that emotion-related information relies more on acoustic and prosodic cues, which are better
preserved before the final contextual layers.

\begin{figure}[!t]
\begin{center}
\begin{tabular}{cc}
\includegraphics[width=0.44\columnwidth]{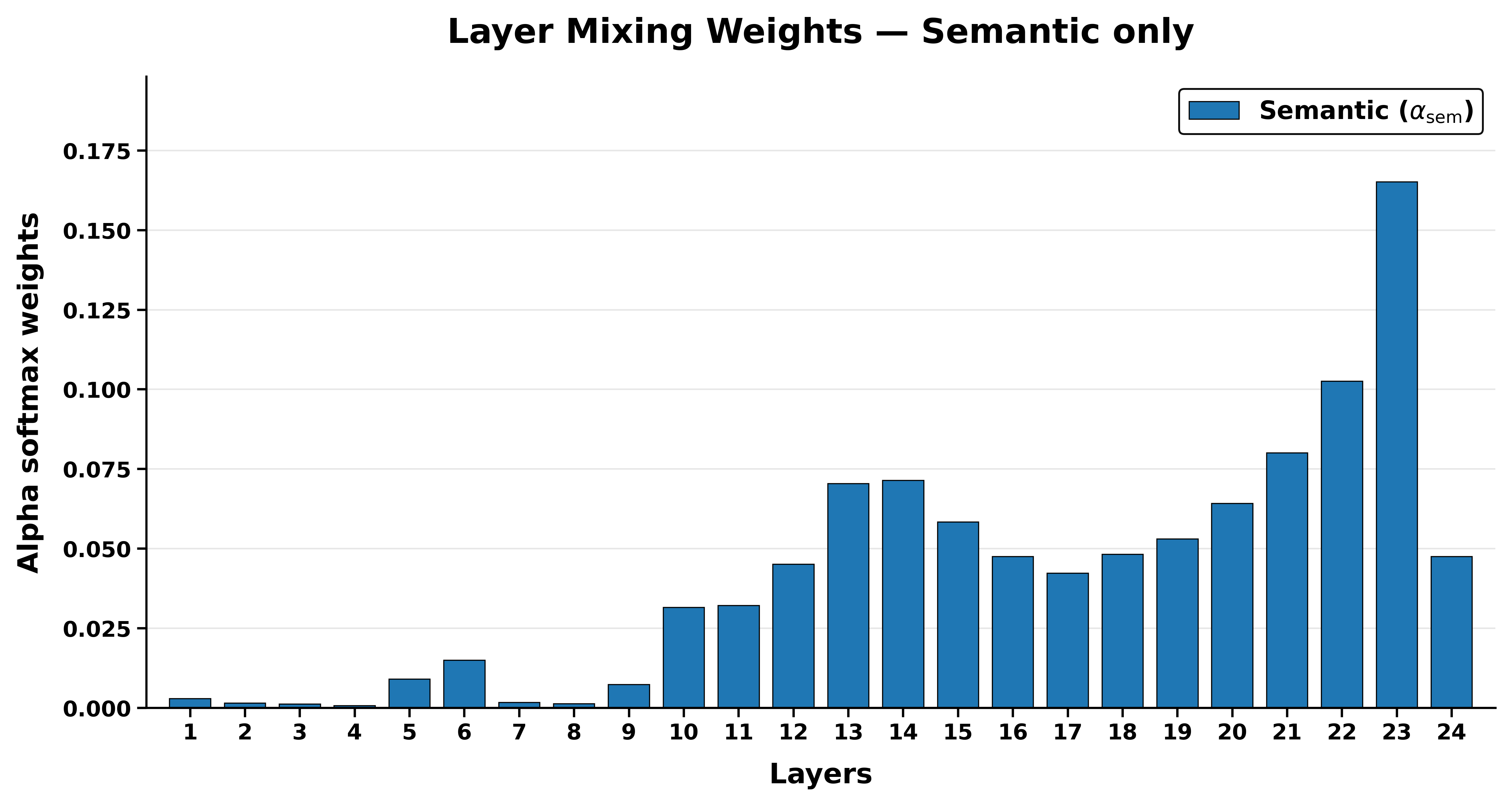}
&
\includegraphics[width=0.44\columnwidth]{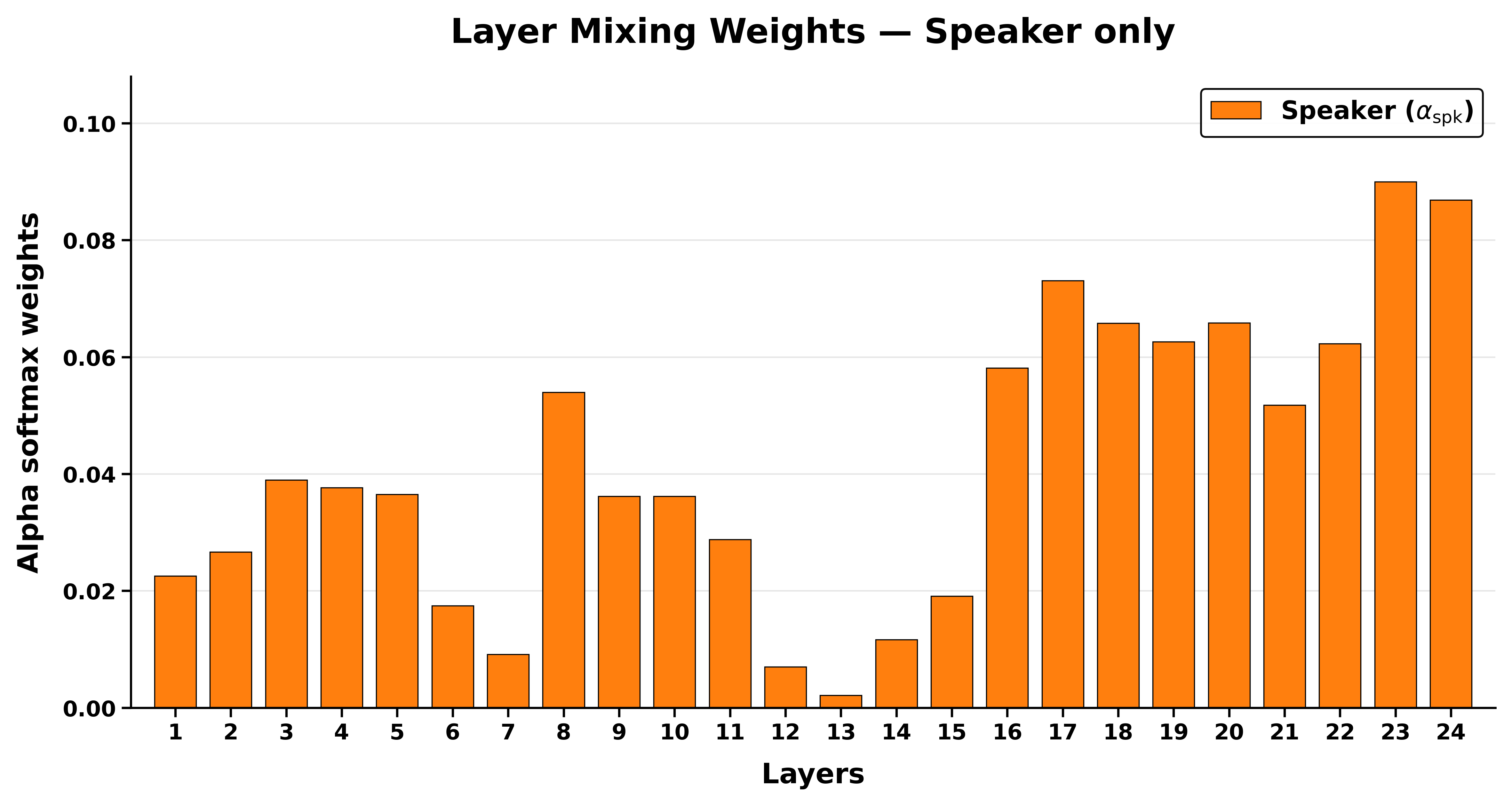}
\\[-0.05cm]
\textbf{(a) SEM-only}
&
\textbf{(b) SPK-only}
\\[0.10cm]
\includegraphics[width=0.44\columnwidth]{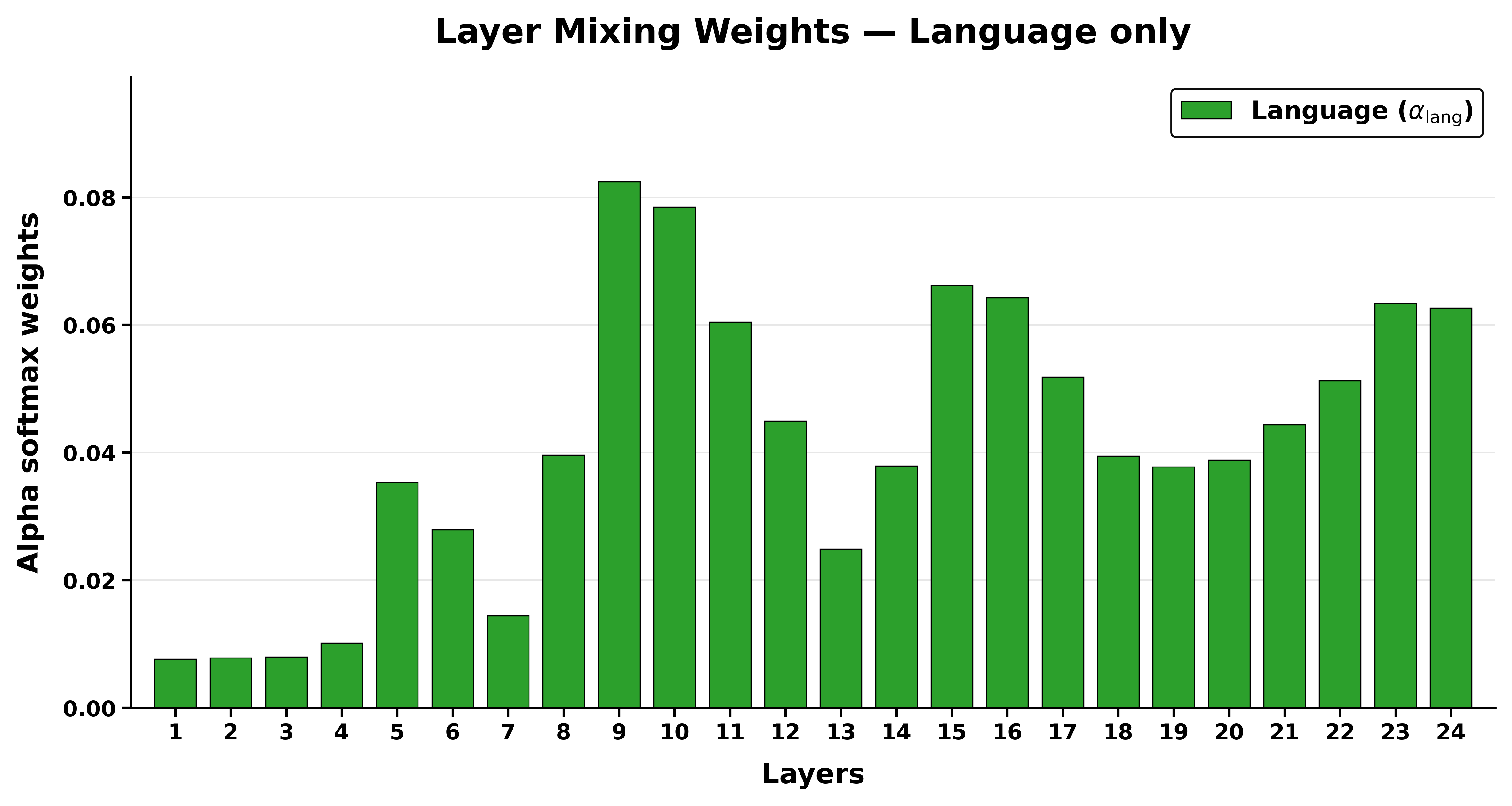}
&
\includegraphics[width=0.44\columnwidth]{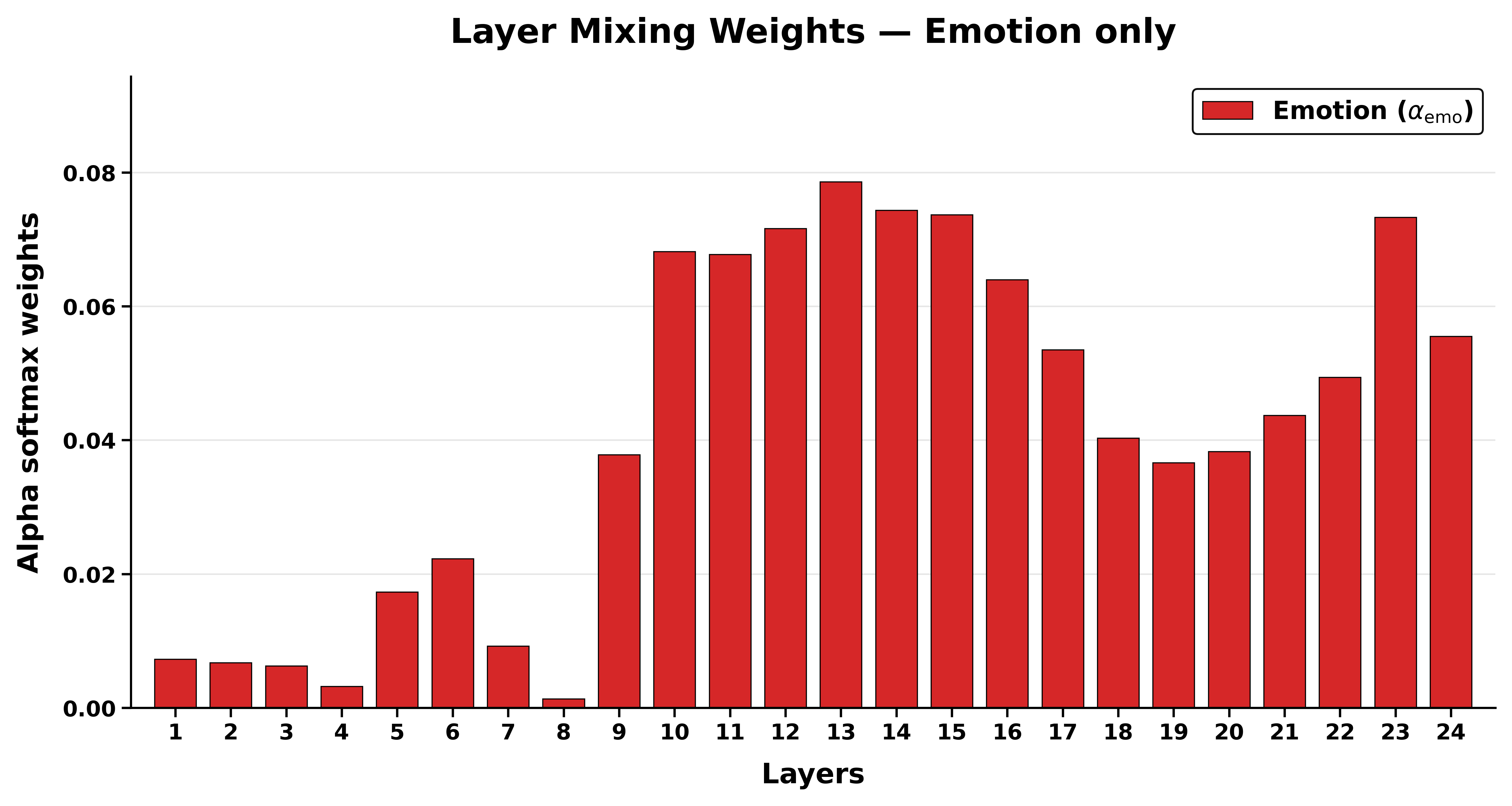}
\\[-0.05cm]
\textbf{(c) LANG-only}
&
\textbf{(d) EMO-only}
\end{tabular}
\end{center}
\caption{Layer-interpolation weights learned by the single-task students. Compared with the Multi-Attribute model, these distributions show that joint training modifies the layer usage of each attribute branch.}
\label{fig:layer_alphas_single_task}
\vspace{-0.4cm}
\end{figure}

Figure~\ref{fig:layer_alphas_single_task} compares the layer-weight distributions of the single-task students. The comparison shows that Multi-Attribute training changes the way information is encoded in the shared encoder. 
For example, the SEM-only student assigns stronger weights to the last layers, whereas the semantic branch of the Multi-Attribute model shifts toward intermediate layers. Similarly, speaker and language information remain broadly distributed, but their layer usage is reorganized when the encoder is jointly optimized with the other objectives.
These observations indicate that the Multi-Attribute model does not simply replicate the single-task representations. 
Instead, joint training redistributes attribute-specific information across the encoder layers.
This depth-wise reorganization helps the shared encoder accommodate semantic, speaker, language, and emotion information within a single model, while preserving competitive performance across the downstream tasks.

\section{Conclusion}

We presented a holistic multi-teacher post-training framework that enables a single self-supervised speech encoder to learn multiple complementary utterance-level speech representations through knowledge distillation. 
By jointly distilling semantic, speaker, language, and emotion representations from four specialized teacher models, the proposed approach replaces several task-specific encoders with a single shared backbone while preserving competitive performance across four independent downstream evaluation tasks.

Despite the heterogeneous nature of the supervision signals and datasets, the proposed model consistently approaches the performance of dedicated single-task students, while even improving semantic retrieval and showing that semantic and language representations can provide complementary information for language identification. 
Beyond the downstream results, the learned layer interpolation weights reveal that different speech attributes naturally emerge at different depths of the shared SSL encoder, providing new insights into the internal organization of speech foundation models.

More broadly, this work suggests that post-training can transform generic SSL speech encoders into holistic representation models capable of supporting multiple speech understanding tasks without requiring dedicated backbones. 
We believe that this direction opens a promising path toward universal speech foundation models producing rich, reusable utterance-level representations for a wide range of downstream applications.

\section{Acknowledgments}


\bibliographystyle{IEEEtran}
\bibliography{mybib}

\end{document}